\documentclass[preprint,12pt]{elsarticle}
\usepackage[utf8]{inputenc}
\usepackage{amsmath}
\usepackage{amssymb}
\usepackage{graphicx}
\usepackage{xcolor}
\usepackage[super]{nth}
\usepackage{geometry}
\usepackage{caption}
\usepackage{subcaption}
\usepackage{float}
\usepackage{hyperref}
\usepackage{multirow}
\usepackage{lineno}

\biboptions{sort&compress}

\title{Discovering a reaction-diffusion model for Alzheimer’s disease by combining PINNs with symbolic regression}

\author{Zhen Zhang\textsuperscript{a}}
\author{Zongren Zou\textsuperscript{a}}
\author{Ellen Kuhl\textsuperscript{b}\corref{cor1}}
\cortext[cor1]{Corresponding author: ekuhl@stanford.edu}
\author{George Em Karniadakis\textsuperscript{a}\corref{cor2}}
\cortext[cor2]{Corresponding author: george\_karniadakis@brown.edu}

\address[a]{Division of Applied Mathematics, Brown University, Providence, RI 02912, USA}
\address[b]{Department of Mechanical Engineering, Stanford University, Stanford, CA 94305, USA}

\begin{document}
\begin{frontmatter}
\begin{abstract}
Misfolded tau proteins play a critical role in the progression and pathology of Alzheimer’s disease. Recent studies suggest that the spatio-temporal pattern of misfolded tau follows a reaction-diffusion type equation. However, the precise mathematical model and parameters that characterize the progression of misfolded protein across the brain remain incompletely understood. Here, we use deep learning and artificial intelligence to discover a mathematical model for the progression of Alzheimer’s disease using longitudinal tau positron emission tomography from the Alzheimer’s Disease Neuroimaging Initiative database. Specifically, we integrate physics informed neural networks (PINNs) and symbolic regression to discover a reaction-diffusion type partial differential equation for tau protein misfolding and spreading. First, we demonstrate the potential of our model and parameter discovery on synthetic data. Then, we apply our method to discover the best model and parameters to explain tau imaging data from 46 individuals who are likely to develop Alzheimer’s disease and 30 healthy controls. Our symbolic regression discovers different misfolding models $f(c)$ for two groups, with a faster misfolding for the Alzheimer’s group, $f(c) = 0.23c^3 - 1.34c^2 + 1.11c$, than for the healthy control group, $f(c) = -c^3 +0.62c^2 + 0.39c$. Our results suggest that PINNs, supplemented by symbolic regression, can discover a reaction-diffusion type model to explain misfolded tau protein concentrations in Alzheimer’s disease. We expect our study to be the starting point for a more holistic analysis to provide image-based technologies for early diagnosis, and ideally early treatment of neurodegeneration in Alzheimer’s disease and possibly other misfolding-protein based neurodegenerative disorders.
\end{abstract}
\end{frontmatter}

% \maketitle
\noindent\textbf{keywords:} Alzheimer's disease, misfolded tau protein, model discovery, PINNs, symbolic regression, uncertainty quantification

\newpage

\section{Introduction}

% Alzheimer's disease

The tau protein plays a critical role in Alzheimer's disease. In healthy brains, tau helps stabilize microtubules to maintain cell structure and transport nutrients. In Alzheimer's disease, tau undergoes abnormal modifications, it misfolds, and forms toxic tangles in the brain. 
The accumulation of misfolded tau contributes to the spread of pathology throughout the brain and correlates with cognitive decline and neurodegeneration. Understanding the spatio-temporal evolution of tau misfolding is vital for developing interventions that target tau pathology and potentially slow down the progression of Alzheimer's disease.

% Diagnosing misfolded tau

Until about ten years ago, the only method to diagnose Alzheimer's disease non-invasively in vivo was cognitive testing to confirm memory loss at the very advanced stages of neurodegeneration. Throughout the past decade, the ability to trace tau protein through positron emission tomography has drastically changed how we can image disease progression across the living brain, non-invasively, at any stage. 
By tracking the accumulation and distribution of misfolded tau over time, these imaging techniques provide insights into disease progression and staging. They allow us to correlated alterations in misfolded tau concentration to cognitive decline, and hold the potential to evaluate therapeutic interventions and monitoring their effectiveness in slowing down disease progression. 
Towards this goal, the Alzheimer's Disease Neuroimaging Initiative (ADNI) database has become the go-to repository of clinical, imaging, and biomarker data from individuals with normal cognition, mild cognitive impairment, and Alzheimer's disease. It contains dozens of freely available, fully annotated, longitudinal scans and enables researchers to analyze and correlate data.

% Modeling tau spreading and misfolding

Computational modeling offers a promising strategy to precisely quantify the accumulation and spreading of misfolded tau proteins across the brain. Mathematical models inspired by reaction-diffusion systems hold significant potential for personalized predictions of disease progression timelines. Previous studies have used cross-sectional positron emission tomography data to calibrate and validate computational models for tau pathology. A simple but efficient model is the Fisher Kolmogorov model combined with network diffusion within the brain's connectome. We have recently shown that this method can be embedded into a hierarchical Bayesian analysis to explain tau protein misfolding and spreading across 83 brain regions from longitudinal neurimaging data of 76 subjects to distinguish Alzheimer's patients and healthy controls \cite{schafer2021bayesian}. However, this method {\it{a priori postulates a mathematical model}} of Fisher-Kolmogorov type for the reaction term and does no allow for alternative functional forms to characterize the complex dynamics of protein misfolding. 

% Here

The objective of this study is to {\it{discover a mathematical model}} for Alzheimer's disease, purely based on clinical data. To this end, we adopt a physics-informed neural network (PINN) \cite{raissi2019physics} and combine it with symbolic regression \cite{Cranmer_Interpretable_Machine_Learning_2023, cranmer2020discovering}. Our group has successfully employed PINNs combined with a Bayesian analysis to study the nonlinear dynamics of disease data \cite{linka2022bayesian} and other groups are now increasingly using PINNs \cite{zou2022neuraluq, kharazmi2021identifiability, yin2023generative} to infer parameters of known models from real-world data. To a priori guarantee physical constraints such as thermodynamic consistency or polyconvexity, recent studies have proposed to hardwire our prior constitutive knowledge into the network input, output, architecture, and activation functions, a strategy that has become known as constitutive artificial neural networks (CANN) \cite{linka2023new}. This type of custom-designed neural networks has been used to discover the model, parameters, and experiment that best explain the behavior of human brain \cite{linka2023automated_brain} and skin \cite{linka2023automated_skin}.

The rest of this paper is organized as follows: in Section \ref{sec:2}, we present the problem formulation and the methodology developed for model discovery (see the PINN method in Section \ref{subsec:pinn} and the symbolic regression method in Section \ref{subsec:symbolic}); in Section \ref{sec:3}, details of the experimental setup and results are discussed, where our approach is first tested on simulated data in Section \ref{subsec:simulated} and then used to discover reaction models for amyloid positive and negative patients from real data in Section \ref{subsec:real}; a discussion regarding the proposed methodology and results is provided in Section \ref{sec:4}.
\section{Methodology}\label{sec:2}

In this section, we formulate the problem and describe the machine learning methodology to discover models for Alzheimer's disease. We focus on the discovery of the reaction model in the reaction-diffusion system, described by the following partial differential equation (PDE):
\begin{equation}\label{eq:problem_1}
    \frac{\partial c}{\partial t} = \nabla \cdot (\textbf{D} \cdot \nabla c) + f(c),
\end{equation}
where $c$ is the concentration of misfolded tau protein, $\textbf{D}$ denotes the heteroscedastic diffusion tensor, and $f: \mathbb{R}\rightarrow \mathbb{R}$ is the reaction model. Different reaction models result in various behaviors of the system, e.g. \cite{fisher1937wave, newell1969finite, segel1969distant, zeldovich1938theory} (see also Table~\ref{tab:KPP_eqs}), and the selection of the appropriate model is governed by the nature of the underlying engineering or life sciences problem. 

Physics-informed neural networks (PINNs) have proven to be powerful tools to solve differential equations as well as infer unknown quantities from data and physical law; see \cite{karniadakis2021physics} for a review and \cite{raissi2019physics, lu2021deepxde, yang2021b, jagtap2020conservative, pang2019fpinns, mao2020physics, cai2021physics, zou2023hydra} for recent developments of PINNs. In this work, PINNs are employed to learn $f(c)$ from data of $c$ and the physics defined in Eq.~\eqref{eq:problem_1}, followed by symbolic regression to determine the analytic expression of $f$. A similar approach was proposed in \cite{podina2022pinn}, in which the PINN and AI Feynman \cite{udrescu2020ai} methods were used to discover differential operators in equations from data. Instead, in this paper, we  focus on the topic of model discovery.

\begin{figure}[H]
\includegraphics[width=\textwidth]{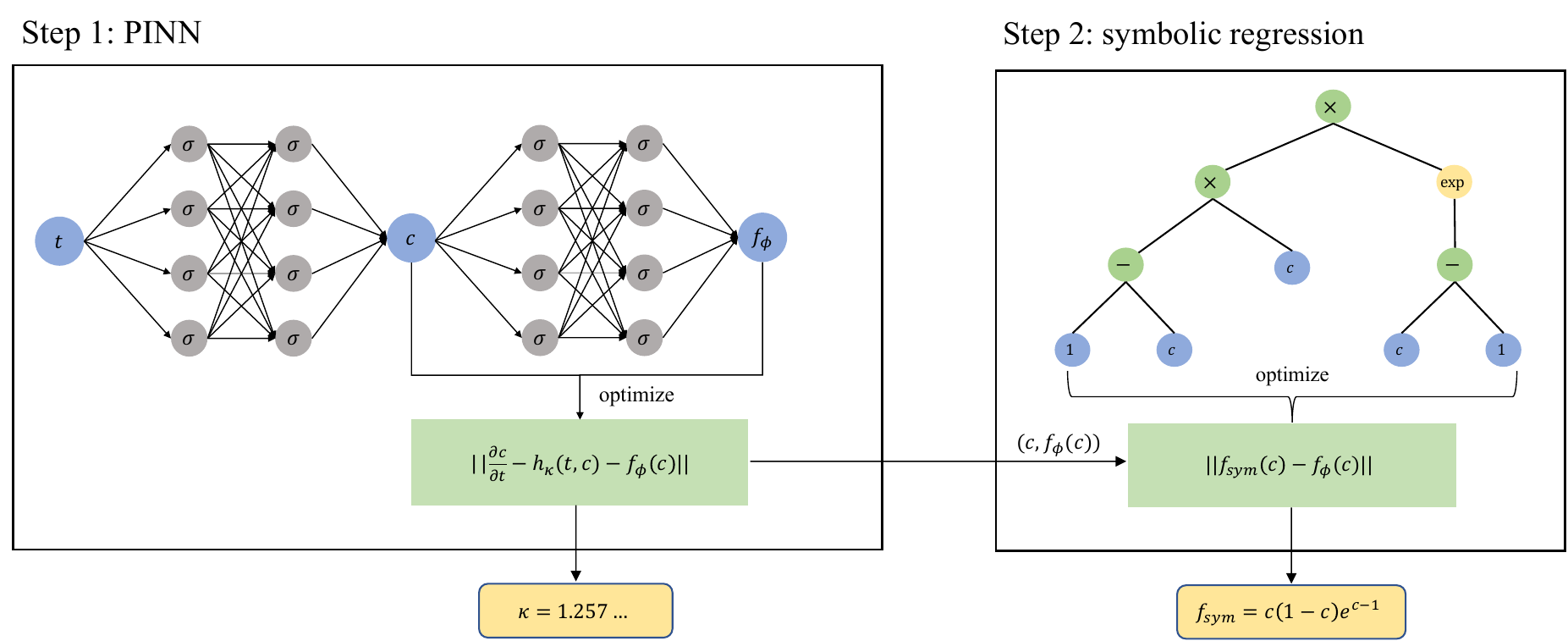}
\caption{\textbf{A schematic view of the approach for model discovery}.
%\textbf{Combination of physics informed neural networks (PINNs) and symbolic regression.} 
The original PINN framework for inferring the diffusion coefficient $\kappa$ and the reaction model $f$ (\textbf{left}) is followed by a symbolic regression step to discover the analytic form of $f$ (\textbf{right}).}
\label{fig:flowchart}
\end{figure}

We follow the same setup as in the literature \cite{schafer2020network, schafer2021bayesian, mcnab2013human} and model the aggregation and propagation of pathological tau in the connectome of the brain by the reaction-diffusion system defined by Eq.~\eqref{eq:problem_1} and discretized on a weighted, undirected graph $G$ \cite{henderson2019connectomics,kuhl2019connectomics}.The nodes of $G$ represent different non-overlapping brain regions and the edges of $G$ are the axonal connections between different regions. 
In this regard, we discretize Eq.~\eqref{eq:problem_1} on $G$, and we recast the PDE as an ODE system:
\begin{equation}\label{eq:problem_2}
    \frac{d c_i}{d t} = h^i_\kappa(t, \mathbf{c}) + f(c_i), i=1,...,N,
\end{equation}
where $\mathbf{c} := [c_1, ..., c_N]^T \in \mathbb{R}^N$ represents the concentration of misfolded tau protein in the $N$ different brain regions, $h^i_\kappa: \mathbb{R}^{N+1} \rightarrow \mathbb{R}$ the diffusion term between different regions, and $f: \mathbb{R} \rightarrow \mathbb{R}$ the local reaction/production model, which characterizes the collective dynamics of protein production, clearance, and conversion from healthy to unhealthy seeds \cite{fornari2019prion}. More details and reasoning of this setup can be found in \cite{schafer2020network, schafer2021bayesian, mcnab2013human} and will be discussed in Section \ref{sec:3}. 
Here the diffusion model, $h_\kappa$, is assumed to be known and parameterized by the diffusion coefficient $\kappa$, while the reaction/production model, $f$, will be discovered. 
Our approach for discovering $f$ is formulated as follows: first infer $\mathbf{c}$, $f$, and $\kappa$ from data of $c$ and the physics defined in Eq.~\ref{eq:problem_2} using PINNs, and then find the analytic expression of $f$ using symbolic regression. 
We remark that in this approach, we use two individual NNs: one for the approximation of $\mathbf{c}$, which takes the time $t$ as input, and one for the approximation of $f$, which takes the misfolded protein concentration $c_i$ as input. Importantly, the data of $\mathbf{c}$, generated either synthetically from simulations or collected longitudinally from medical images, are only used in the first step.
A schematic view of the workflow is illustrated in Fig.~\ref{fig:flowchart}, assuming a homogeneous distribution with $N=1$ and no diffusion. A pedagogical example on discovering the Kraichanan-Orszag dynamical system \cite{wan2006multi, zou2022neuraluq, chen2023leveraging} and a tutorial of this approach can be found in Appendix \ref{appendix:A} for better understanding. 

\subsection{\normalfont{\textbf{Physics-informed neural networks (PINNs)}}}\label{subsec:pinn}

The PINN method, originally proposed by Raissi et. al. in \cite{raissi2019physics}, addresses ODE/PDE problems by deploying neural networks (NNs) as surrogate models for quantities of interest, constructing a loss function with respect to the ODE/PDE used to describe the physics, and optimizing NN parameters such that the loss function is minimized (see \cite{raissi2019physics, lu2021deepxde} for more details). In our case, $\mathbf{c}$ and $f$ are approximated with NNs, denoted by $\mathbf{c}_{\theta}$ and $f_{\phi}$, respectively, where $\theta$ and $\phi$ denote NN parameters, e.g., weights and biases. 
Let $\mathcal{T}_D$ denote the set of $t$ on which data of $\mathbf{c}$ is (partially) available and $\mathcal{T}_R$ denote the set of $t$ on which residuals of the equation are computed. Gradient-descent based methods, e.g., Adam \cite{kingma2014adam} and L-BFGS \cite{zhu1997algorithm}, can be employed to minimize the following loss function:
\begin{equation}\label{eq:loss}
    \mathcal{L}(\Theta) = \mathcal{L}_{data}(\theta) + \mathcal{L}_{res}(\theta, \phi, \kappa) + \mathcal{L}_{aux}(\theta, \phi, \kappa)
\end{equation}
with respect to $\Theta := \{\theta, \phi, \kappa\}$,
where 
\begin{equation}
    \mathcal{L}_{data}(\theta) = \frac{\sum_{t\in \mathcal{T}_D}||\mathbf{c}(t) - \mathbf{c}_{\theta}(t)||_2^2}{|\mathcal{T}_D|},
\end{equation}
\begin{equation}
    \mathcal{L}_{res}(\theta, \phi, \kappa) = \frac{\sum_{t\in\mathcal{T}_R}||\frac{d \mathbf{c}_{\theta}}{d t}(t) - h_\kappa(\mathbf{c}_\theta)-f_\phi(\mathbf{c}_\theta)||_2^2}{|\mathcal{T}_R|},
\end{equation}
and $\mathcal{L}_{aux}$ is the auxiliary loss that can be seen as additional physical information. More details will be discussed in Section \ref{sec:3} and in Eq.~\eqref{eq:auxiliary_loss}. 

It is possible to enforce boundary conditions on $f$ or $\mathbf{c}$ \cite{lu2021deepxde, huang2023solving, berrone2022enforcing}. For example, for $f$ one can set 
\begin{equation}
    f_\phi(\mathbf{c}) = G(\mathbf{c}) + D(\mathbf{c})\odot g_\phi(\mathbf{c}),
\end{equation}
where $D(\mathbf{c})$ represents the distance from $\mathbf{c}$ to the boundary, $G$ is a function that matches $f$ on the boundary, while $g_\phi$ is a fully-connected neural network.

\subsection{\normalfont{\textbf{Symbolic regression}}}\label{subsec:symbolic}
For symbolic regression, we adopt the open-source software PySR \cite{Cranmer_Interpretable_Machine_Learning_2023}, which  has a configurable Python interface built on the efficient Julia backend \textit{SymbolicRegression.jl}. The underlying algorithm for PySR involves tree search and regularized evolution. In our case of study, we use PySR to distill knowledge from $f_\phi$ to obtain a symbolic expression $f_{sym}$. 

\subsubsection*{\normalfont{\textbf{Evaluation Metric}}}
We use \textit{score} defined in \cite{cranmer2020discovering} as one of our evaluation metrics for the correctness of an expression,
\begin{equation*}
    score = -\Delta \log(\text{MAE}) / \Delta C,
\end{equation*}
where MAE is the mean absolute error between the prediction and the data, $C$ refers to the complexity of the expression, and $\Delta$ denotes local change \cite{cranmer2020discovering}. Higher \textit{score} means that with a slightly lower complexity, MAE of the symbolic regression model becomes much larger. A model with low loss and high score is preferred, so we pick the expression with the highest score among those whose loss is lower than 1.5 times the loss of the most accurate model as our final candidate. 

When multiple simulations are conducted in the PINN inference stage, one may obtain multiple functions $f_{\phi}$. The score metric can be used to pick one $f_{sym}$ for each $f_\phi$. Then, in order to compare $f_{sym}$ obtained from different $f_{\phi}$, we use the projection error as another evaluation metric. We first solve Eq.~\eqref{eq:problem_2} with the inferred parameter $\kappa$ from the PINN in step 1 and the discovered model $f_{sym}$ from the symbolic regression in step 2 to obtain the solution, denoted by $\mathbf{c}_{proj}$, and then we evaluate $\mathbf{c}_{proj}$ on $t\in\mathcal{T}_D$. Specifically, the projection error is defined by
 \begin{equation}
    \text{projection error} = \frac{\sum_{t\in \mathcal{T}_D}||\mathbf{c}(t) - \mathbf{c}_{proj}(t)||_2^2}{|\mathcal{T}_D|}.
\end{equation}
The candidate with the lowest projection error is the one that fits the data best.

\section{Results}\label{sec:3}

In this section, we discuss details of the experimental setup and discovery of the reaction model of the reaction-diffusion system. We first test the approach with synthetic data in Section \ref{subsec:simulated}, in which we use data from simulations of Eq.~\eqref{eq:problem_2} with four different reaction models. 
Then we move to real data in Section \ref{subsec:real}, where two models are discovered using our approach with uncertainty. Specifically for real data, we assume that all individuals within each group, amyloid positive and amyloid negative, share the same type of reaction model but personalized with different reaction rates. 
A deep ensemble method for PINNs \cite{lakshminarayanan2017simple, zou2022neuraluq, psaros2023uncertainty} is employed for uncertainty quantification.

As mentioned in Section \ref{sec:2}, to apply the reaction-diffusion equation to our brain-related problems, we follow the same setup as in \cite{schafer2021bayesian} and discretize Eq.~\eqref{eq:problem_1} on a graph $G$, which comes from the Budapest Reference Connectome v3.0 \cite{szalkai2017parameterizable} and the Human Connectome Project \cite{mcnab2013human}; see \cite{schafer2021bayesian} for details on how to construct $G$. There are in total $N=83$ nodes in the graph $G$, representing 83 considered cortical and subcortical brain regions. In this work, we adopt the graph $G$ and discretization from \cite{schafer2021bayesian} and Eq.~\eqref{eq:problem_1} is discretized as follows:
\begin{equation}\label{eq:discretized}
    \frac{dc_i}{dt} = -\kappa \sum_{j=1}^{N}L_{ij}c_j+\alpha f(c_i), i=1,...,N,
\end{equation}
where $c_i$ is the concentration of tau protein in brain region $i$, $\kappa$ determines the transport rate of misfolded protein between regions, $(L_{ij})$ is the graph Laplacian, representing the connectivity of the graph, and $\alpha$ denotes the reaction rate; $\kappa$ and $\alpha$ are assumed to be specific to each individual but shared across all regions. We note that this corresponds to Eq.~\eqref{eq:problem_2} with $h^i_\kappa(\mathbf{c}) = -\kappa \sum_{j=1}^{N}L_{ij}c_j$.

Unlike previous studies \cite{fornari2019prion, schafer2021bayesian}, which assume a single specific reaction model, $f(c) = c(1-c)$, herein we aim to identify the analytical form of the function $f$ from tau concentration data of 76 subjects from the Alzheimer's Disease Neuroimaging Initiative (ADNI) database. 

The intuition behind this is as follows: Assuming the same reaction model for all amyloid positive and negative subjects might be too constrained and have a negative effect when the fitting of data. However, we found empirically that requirements from the Fisher equation \cite{fisher1937wave} and the KPP equation \cite{kolmogorov1937study} on the reaction term were necessary to provide better and more realistic results. As a generalized form of the Fisher equation, the KPP equation requires that the reaction term $f$ has the following properties:
\begin{linenomath}
\begin{subequations}
\begin{align}
    &f(0) = f(1) = 0, f(c) > 0, \label{eq: kpp_0} \\
    &f'(c) < f'(0), \forall c \in [0,1] \label{eq: kpp_1},
\end{align}
\end{subequations}
\end{linenomath}
which can be considered as the prior knowledge or constraints of the to-be-discovered model. 
To satisfy constraints in Eqs.~\eqref{eq: kpp_0} and \eqref{eq: kpp_1}, we first set 
\begin{equation}\label{eq:nn_constraint}
    \Tilde{f}_{\phi}(c) = c(1-c)e^{g_\phi(c)}
\end{equation} to enforce \eqref{eq: kpp_0}, where $g_\phi$ is a standard fully-connected NN parameterized by $\phi$. Then $\Tilde{f}_{\phi}$ is normalized by setting
\begin{equation}
    f_\phi(c) = \frac{\Tilde{f}_{\phi}(c)}{4 \max_{x\in [0,1]}\Tilde{f}_{\phi}(x)}
\end{equation}
as our final parameterization of $f$ to ensure the inferred reaction term does not degenerate to 0. We introduce
\begin{equation}\label{eq:auxiliary_loss}
    \mathcal{L}_{aux} = \frac{\sum_{c\in \mathcal{C}}||\max(0, \alpha f_{\phi}'(c) - \alpha f_{\phi}'(0))||_1}{|\mathcal{C}|}, 
\end{equation}
in \eqref{eq:loss} as the auxiliary loss term to softly enforce condition \eqref{eq: kpp_1}, which guarantees a travelling wave solution to the Fisher-Kolmogorov equation. In this paper we set $\mathcal{C} = \{ 0, 0.01, 0.02, \cdots, 1\}$. Another reason for introducing this additional regularization term is to promote smoothness of the inferred reaction term. Additional types of KPP equations, including the Newell-Whitehead-Segel equation and the Zeldovich-Frank-Kamenetskii equation, are summarized in Table~\ref{tab:KPP_eqs}. 

\begin{table}[H]
\centering
\begin{tabular}{|c|c|c|c|c|}
\hline
Group & Equation & General form & Reference & Our case \\ \hline
1 & Fisher & $kc(1-c)$ & \cite{fisher1937wave} & $c(1-c)$ \\ \hline
2 & \multirow{2}{*}{Newell-Whitehead-Segel} & \multirow{2}{*}{$kc(1-c^q)$} & \multirow{2}{*}{\cite{newell1969finite, segel1969distant}} & $\frac{3\sqrt{3}}{8}c(1-c^2)$ \\ \cline{1-1} \cline{5-5} 3 & & & & $\frac{2^{2/3}}{3}c(1-c^3)$ \\ \hline
4 & Zeldovich-Frank-Kamenetskii & $kc(1-c)e^{\beta(c-1)}$ & \cite{zeldovich1938theory} & $\frac{\sqrt{5}+2}{4}c(1-c)e^{c-1-\frac{\sqrt{5}-3}{2}}$ \\ \hline
\end{tabular}
\hfill
\caption{\textbf{Reaction term $f$ in KPP equations.} We stratify the subjects into 4 groups with the assumption that subjects in each group share the same reaction term $f(c)$ in Section \ref{subsec:simulated}. Diverse reaction terms are associated with distinct KPP equations, which are in turn indicative of particular biological characteristics. We normalize $f$ to make its maximum value  $\frac{1}{4}$ in all groups.}
\label{tab:KPP_eqs}
\end{table}

\subsection{\normalfont{\textbf{Synthetic data}}}\label{subsec:simulated}
\subsubsection*{\normalfont{\textbf{Data preparation}}}
In this study, we conducted a simulation of tau concentration for a sample of 76 subjects with varying initial conditions, parameters, and reaction terms, employing Equation~\eqref{eq:discretized}. The subjects were stratified into four groups of equal size, with differing reaction terms across groups, as detailed in Table~\ref{tab:KPP_eqs}. The parameters $\kappa$, $\alpha_i$, and $\alpha_{ij}$ were assumed to follow probability distributions of $\text{BoundNormal}(1, 0.5^2)$, $\mathcal{N}(0.6, 0.1^2)$, and $\mathcal{N}(\alpha_i, 0.2^2)$, respectively, where $\alpha_{ij}$ denotes the $j$-th subject in the $i$-th group. The initial tau concentration for the $i$-th node, denoted by $c_i(0)$, was sampled from a normal distribution with equivalent mean and variance as the real data. The connectivity matrix, denoted by $L$, was set to be identical to the real data. The tau concentrations $c(t_k)$ are sampled at $t_k = k$ for $k = 0, 1, 2$ years.

\subsubsection*{\normalfont{\textbf{Parameter and function identification}}}
\begin{figure}[!htbp]
\includegraphics[width=\textwidth]{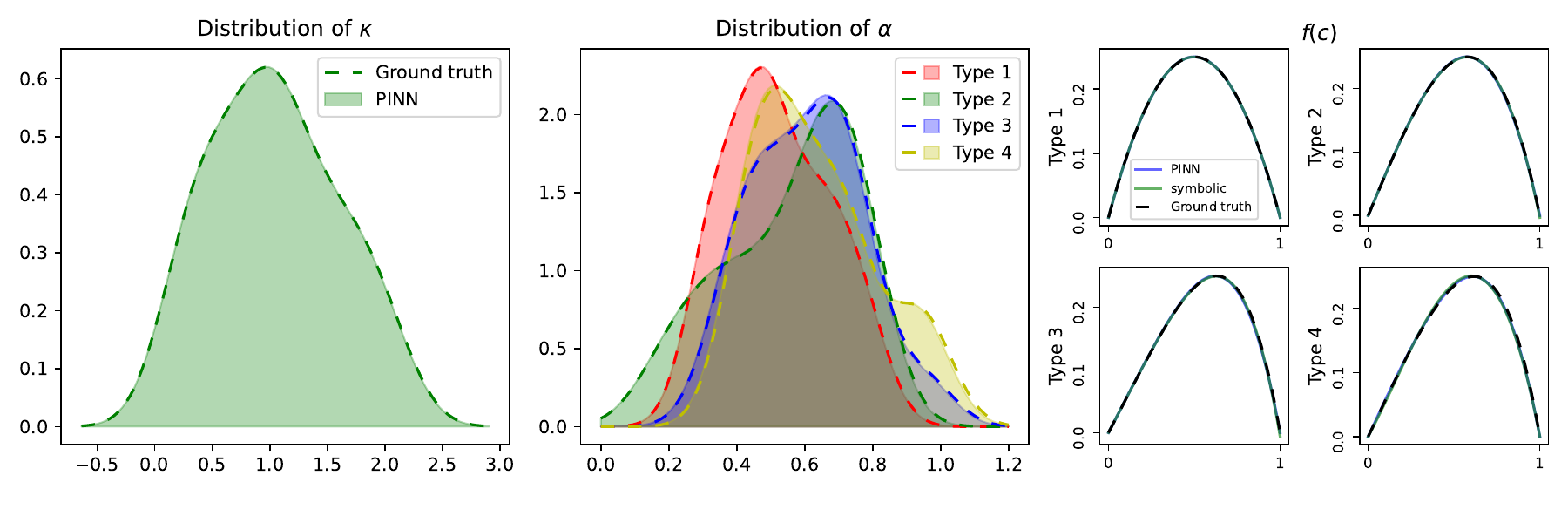}
\caption{\textbf{Parameter and function identification results on synthetic data.} Inference results for transport rate (\textbf{left}) and local production rate (\textbf{middle}), as well as the reaction term for each group (\textbf{right}). PINNs accurately identify the distribution of the transport rate at the global level and the distribution of $\alpha$ in each group. The neural network inference result, $f_{\phi}$, as well as the symbolic regression approximation, $f_{sym}$ agree with the ground truth $f$ for all four cases.}
\label{fig:params}
\end{figure}

\begin{figure}[!htbp]
\includegraphics[width=\textwidth]{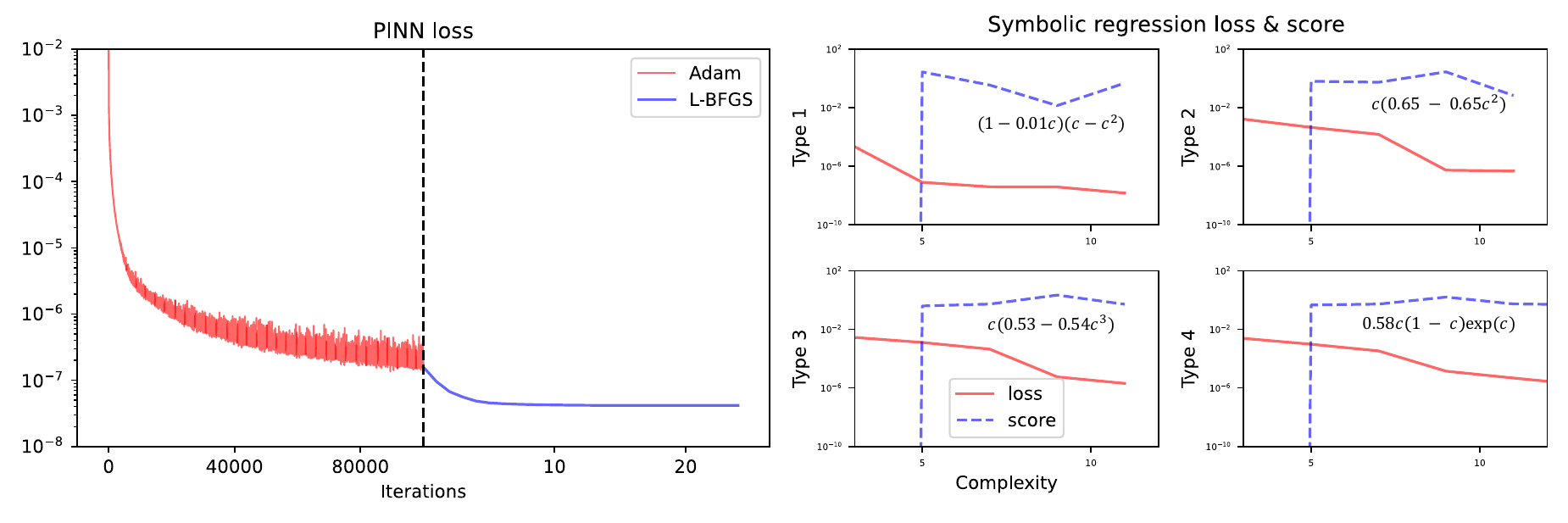}
\caption{\textbf{Losses and evaluation metrics for PINNs and symbolic regression on synthetic data.} \textbf{(Left)} Training loss $\mathcal{L} = \mathcal{L}_{data} +\mathcal{L}_{res}$ for PINN. L-BFGS is applied following Adam to ensure convergence. \textbf{(right)} Loss and score for the symbolic regression model at each complexity level. The inferred $f_{sym}$ is highlighted. It aligns with the correct reaction function, and the plateau in loss suggests minimal loss improvement by increasing the expression complexity.}
\label{fig:statistics_sim}
\end{figure}

The initial step in the present study involves the determination of undetermined parameters and remainder terms, specifically $\kappa$ and $\alpha f(c)$, through the application of PINNs. We set $\alpha$ and $\kappa$ as subject-specific learnable parameters and approximate $f(c)$ with a neural network $f_{\phi}$, which is hard constrained as indicated in \eqref{eq:nn_constraint}. 
The training loss is shown in the left portion of Figure~\ref{fig:statistics_sim}. The outcomes of the identification process are displayed in Figure~\ref{fig:params}. The left portion of the figure displays the inferred and actual distributions of the transport rate, $\kappa$, at the global level, utilizing the Gaussian kernel density estimator \cite{parzen1962estimation}. The congruence between the two distributions signifies that $\kappa$ can be accurately identified. In the middle of the figure, the inferred and actual distributions of the local production rate, $\kappa$, at the group level are illustrated. Notably, the PINNs method successfully captures the appropriate distribution of the growth rate $\alpha$ in each group, regardless of the corresponding reaction term. The right portion of the figure presents the inferred and actual reaction terms for each group, demonstrating that the neural network inference result, $f_{\phi}$, concurs with the ground truth $f$ for all four cases. The effect of hard constraining $f_{\phi}$ is discussed in Appendix \ref{appendix:sample_size}, where we found that without the hard constraint, $f_{\phi}$ eventually deviates from the ground truth on regions where there is no data for $c$, which affects the prediction capability of the entire framework. This is illustrated in Figure~\ref{fig:params_appendix} that we need data up to $t = 6$, meaning medical images for consecutive six years, if there is no constraint on the boundary values.

In the second stage of our analysis, we utilize the symbolic regression package PySR to discover closed-form expressions for the four functions $f_{\phi}$. To achieve this, we minimize the mean squared error (MSE) as the target function. Each $f_{\phi}$ undergoes 100 iterations of simulation. The binary operators are chosen to be addition, subtraction, and multiplication, the unary operators are exponential and reciprocal, and the complexities of unary operators are set to be 3 and 3. PySR generates analytical expressions, evaluates their corresponding losses (MSE), and scores at every complexity level. This information is presented in the right portion of Figure~\ref{fig:statistics_sim}. Subsequently, for each subject, we follow the procedures outlined in Section \ref{subsec:symbolic} to select a final candidate expression $f_{sym}$ as the associated reaction term. We present the four chosen expressions along with their respective scores highlighted with a black circle. Notably, the discovered $f_{sym}$ aligns with the correct reaction function. Furthermore, a plateau in loss is observed after selecting the best candidate, suggesting that increasing the complexity of the discovered function beyond this point would only result in a minimal improvement of the loss.

\subsubsection*{\normalfont{\textbf{Projection of tau concentration}}}
\begin{figure}[!htbp]
\centering
\begin{subfigure}{\textwidth}
    \centering \includegraphics[width=0.99\textwidth]{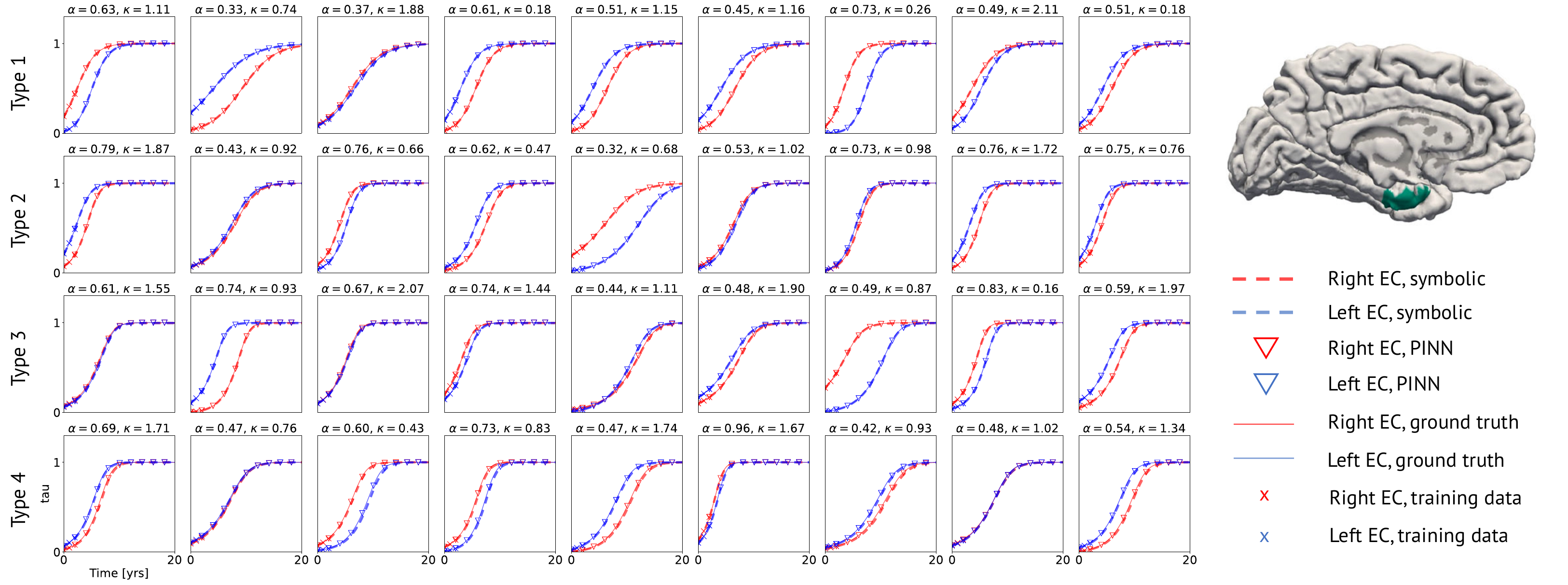}
\end{subfigure} \\
\begin{subfigure}{\textwidth}
    \centering \includegraphics[width=0.99\textwidth]{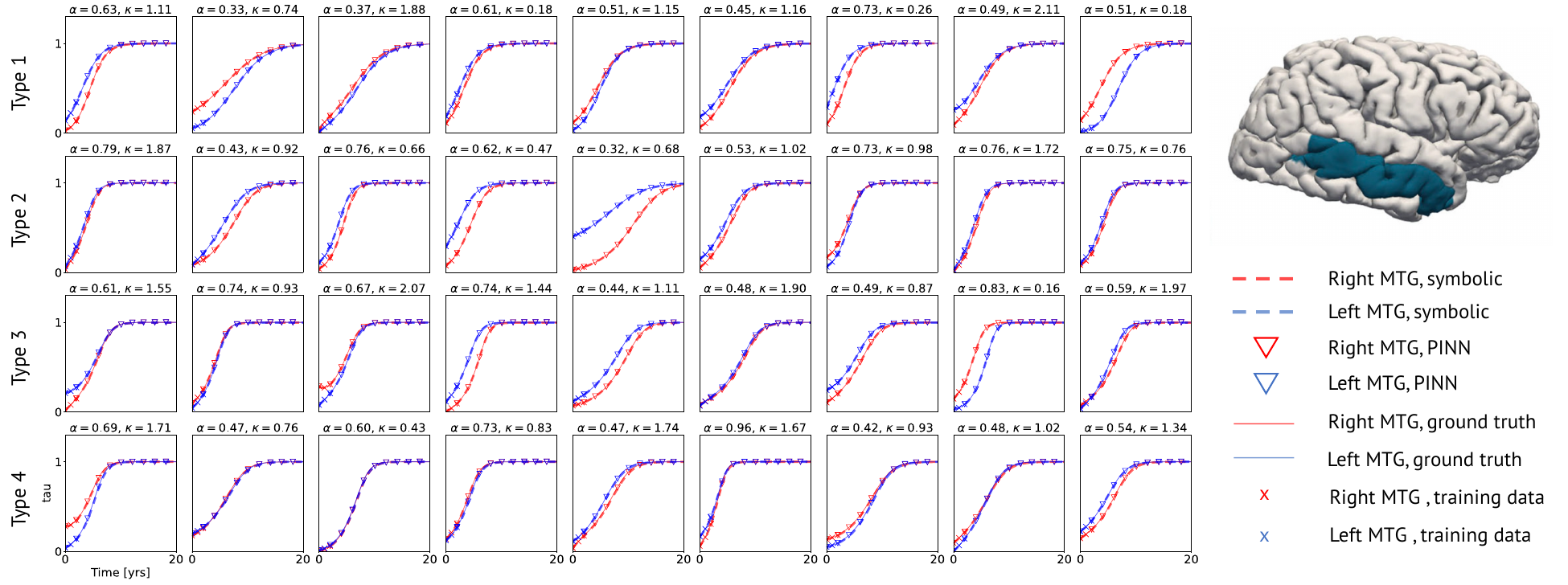}
\end{subfigure} \\
\begin{subfigure}{\textwidth}
    \centering \includegraphics[width=0.99\textwidth]{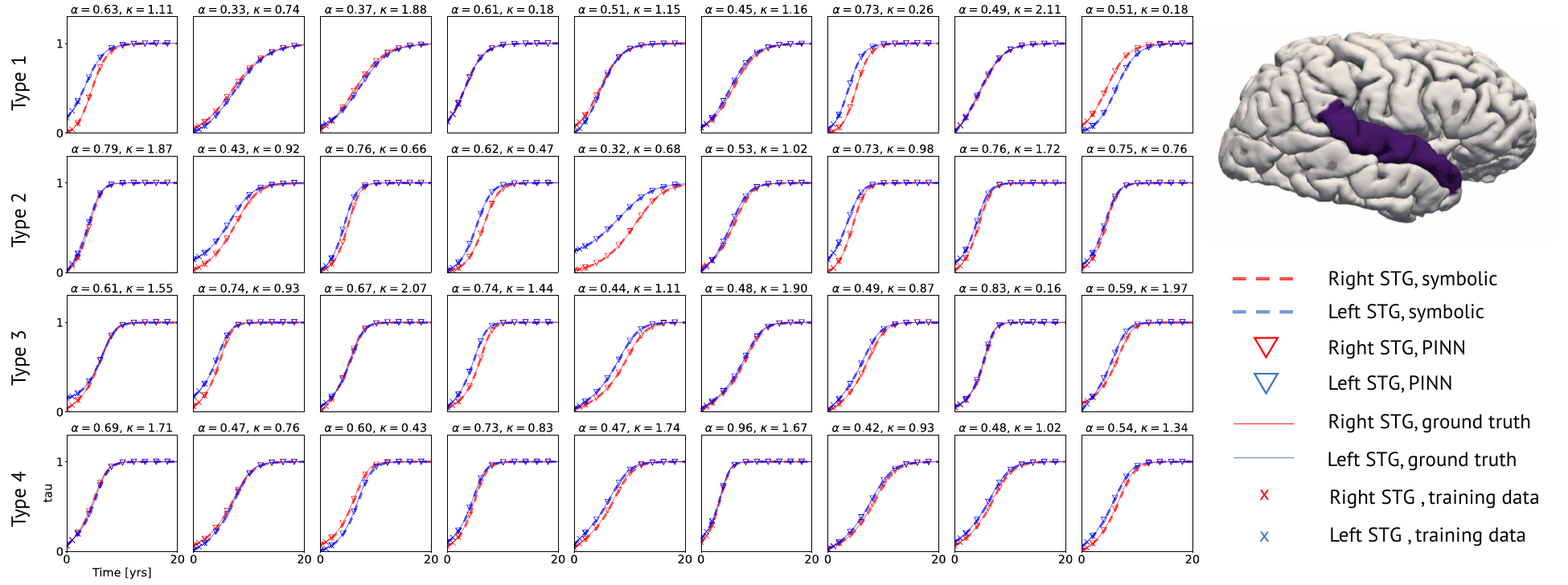}
\end{subfigure} 
\caption{\textbf{Prediction of tau concentration.} Evolution of misfolded tau protein in entorhinal cortex (EC), middle temporal gyrus (MTG), and superior temporal gyrus (STG) throughout 20 years based on synthetic data from the first two years.}
\label{fig:projection}
\end{figure}

In the last stage of our analysis, after we have identified the unknown parameters $\kappa$ and $\alpha$ and the unknown functions ($f_{\phi}$ and $f_{sym}$), we substitute them back into the ODE \eqref{eq:discretized} and solve the equation up to $t = 20$ with the built-in Python method \emph{scipy.integrate.odeint} to examine the predictability of our models. The projections of the tau concentration over 20 years after the first PET scan in 36 subjects (nine from each group) and three different brain regions, namely entorhinal cortex (EC), middle temporal gyrus (MTG), and superior temporal gyrus (STG), are illustrated in Fig.~\ref{fig:projection}. The projection outcome achieved by PINNs, alongside the symbolic regression model, demonstrates a high degree of concordance with the ground truth solution. Moreover, we present the deduced values of $\alpha$ and $\kappa$ for the 36 subjects, which are located at the top of every subfigure. These inferred values are also in agreement with the preset parameters.

\subsection{\normalfont{\textbf{Real data}}}\label{subsec:real}
\subsubsection*{\normalfont{\textbf{Data preparation}}}
We utilize preprocessed tau concentration data directly from the literature \cite{schafer2021bayesian}. In total 76 subjects are classified into two groups: 46 are identified as amyloid positive ($A\beta^+$) meaning their mean amyloid concentration exceeds a certain level, and 30 are identified as amyloid negative ($A\beta^-$). For each subject, three data points are provided, which are on average one year ($1.07\pm0.31$) apart.

\subsubsection*{\normalfont{\textbf{Parameter and function identification with uncertainty quantification}}}
\begin{figure}[!htbp]
\includegraphics[width=\textwidth]{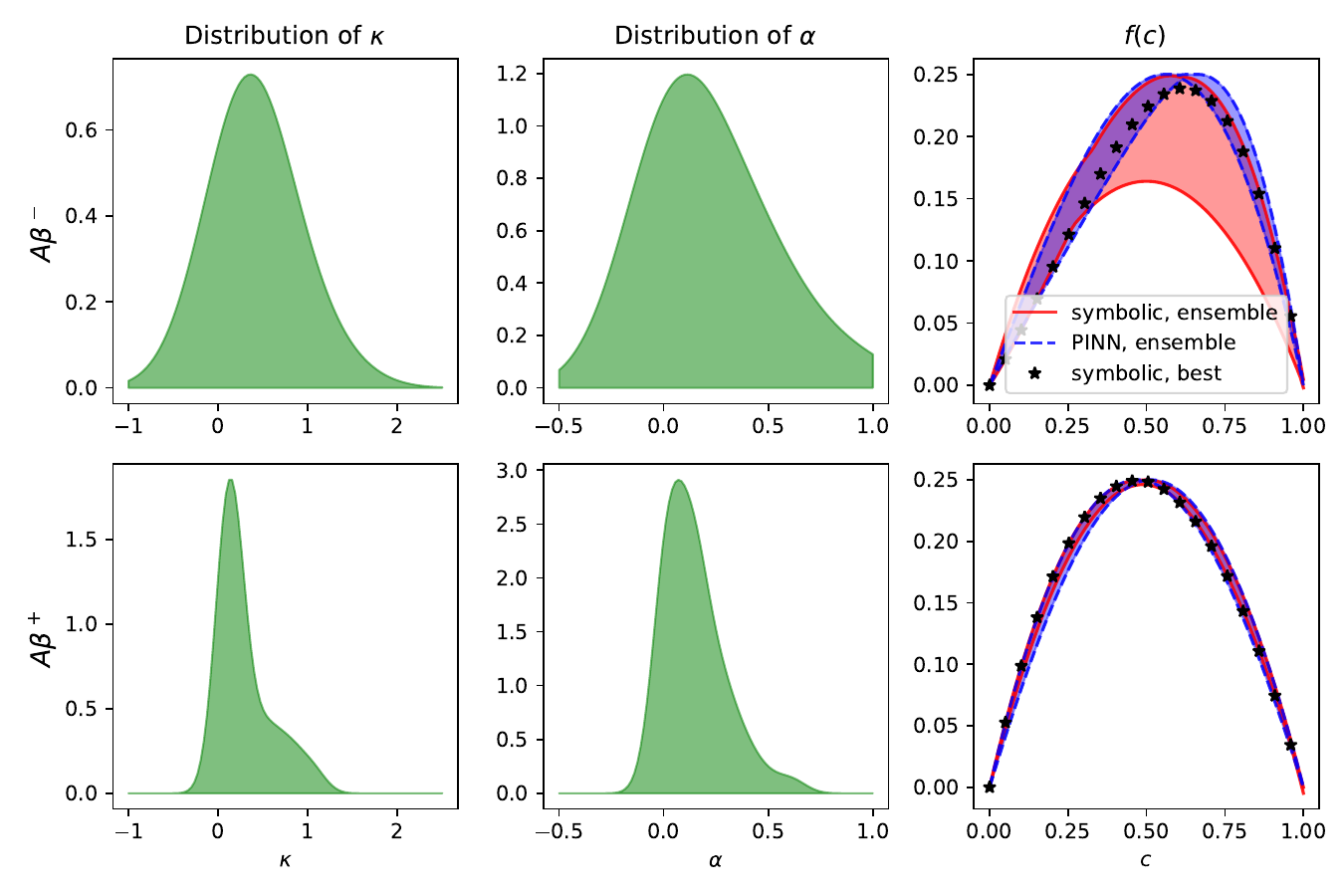}
\caption{\textbf{Parameter and function identification results on real data.} We present the inference results for the transport rate $\kappa$ and local production rate $\alpha$, alongside the reaction term for both the positive and negative groups. (\textbf{Left, Middle}) To present the obtained results, we only visualized the (population-level) distribution of $\alpha$ and $\kappa$ in the simulation with the lowest projection error, as it is not appropriate to aggregate the parameter distribution corresponding to different $f$s. (\textbf{Right}) We plot the ensemble of inferred $f_{sym}$ and $f_{\phi}$, with shaded region representing the minimum and maximum value of $f$s in all 10 simulations. We highlight the $f_{sym}$ which corresponds to the simulation with the lowest projection error with stars.}
\label{fig:params_real}
\end{figure}

Observational noise in the real-life dataset could significantly increase the training difficulty of neural network models as well as complicate the optimization landscape. For such highly nonconvex optimization problem, the parameter initialization becomes more important. Different neural network initialization could lead to drastically different solutions and randomness in the learned parameter set. Even though sometimes it is difficult for the model to identify global minima for every initialization, most modern optimization techniques can handle local minima and ultimately find a parameter set close to the the global minimum. We can interpret the randomness in the inferred parameters under independent simulations as model uncertainty, and such uncertainty quantification method is commonly used in machine learning due to its simplicity yet effectiveness \cite{lakshminarayanan2017simple, zou2022neuraluq, psaros2023uncertainty} and often referred to as deep ensemble. In this work, we conduct ten independent simulations in the PINN inference stage to obtain ten different parameter sets of $\alpha$ and $\kappa$ for each subject, as well as ten different functions $f_\phi^+$ and $f_\phi^-$ for the amyloid positive and negative groups. The distribution of the personalized parameters $\kappa$ and $\alpha$ is shown in the left two columns of Fig.~\ref{fig:params_real}. Note that we only present the distribution for the best of all ten simulations, i.e., the simulation with lowest projection error, since it is not appropriate to aggregate parameter distributions corresponding to different $f$. On the rightmost column of Fig.~\ref{fig:params_real}, we show the ensemble of ten $f_\phi^-$, $f_\phi^+$, as well as $f_{sym}^-$, $f_{sym}^+$. It can be seen that in ten independent simulations, the reaction term clustered into two modes depending on whether the subject is from the amyloid positive or negative group. Interestingly, the variation of the discovered function $f_{sym}$ appears to be larger for the negative group.

% {\color{blue} TODO: add description of uncertainty quantification in scientific machine learning and model discovery \cite{schafer2021bayesian, linka2022bayesian, linka2023automated, linka2023new}...} Deep learning framework and using neural networks often involve randomness in modeling, training, and inference. Hence awareness and quantification of uncertainty in scientific machine learning are crucial in obtaining accurate and trustworthy predictions \cite{zou2022neuraluq, zou2023hydra, psaros2023uncertainty}.

Table~\ref{tab:real_data_exp} summarizes the discovered reaction terms for each simulation together with their projection errors, as defined in Section~\ref{subsec:symbolic}. For each group, we highlight the expression with lowest projection error. Stricingly, our symbolic regression model discovers different misfolding models $f(c)$ for the two groups, with a steeper increase for the Alzheimer’s group, $f(c) = 0.23c^3 - 1.34c^2 + 1.11c$, than for the healthy control group, $f(c) = -c^3 +0.62c^2 + 0.39c$.
\begin{table}[!htbp]
%\footnotesize
\centering
\begin{tabular}{|l|c|c|c|c|}
\hline
\multirow{11}{*}{$A\beta^-$} & $\kappa$       & $\alpha$       & $f_{sym}$                   & Projection error                   \\ \cline{2-5} 
                             & $0.49\pm 0.51$ & $0.24\pm 0.41$ & $-0.64c^3 + 0.64c$          & 7.19E-04                           \\ \cline{2-5} 
                             & $0.50\pm 0.57$ & $0.30\pm 0.48$ & $-c^3 + 0.62c^2 + 0.39c$    & 4.94E-04                           \\ \cline{2-5} 
                             & $0.52\pm 0.62$ & $0.30\pm 0.47$ & $-0.86c^2 + 0.86c$          & 2.09E-03                           \\ \cline{2-5} 
                             & $0.60\pm 1.02$ & $0.28\pm 0.45$ & $-c^3 + 0.59c^2 + 0.41c$    & 4.95E-04                           \\ \cline{2-5} 
                             & $0.49\pm 0.48$ & $0.29\pm 0.28$ & $-0.65c^2 + 0.65c$          & 6.45E-04                           \\ \cline{2-5} 
                             & $0.49\pm 0.50$ & $0.24\pm 0.40$ & $-0.66c^2 + 0.66c$          & 6.00E-04                           \\ \cline{2-5} 
                             & $0.51\pm 0.59$ & $0.24\pm 0.40$ & $-0.64c^2 + 0.64c$          & 7.08E-04                           \\ \cline{2-5} 
                             & $0.57\pm 0.88$ & $0.29\pm 0.45$ & $-c^3 + 0.62c^2 + 0.39c$    & \textbf{4.80E-04} \\ \cline{2-5} 
                             & $0.51\pm 0.62$ & $0.27\pm 0.45$ & $-c^3 + 0.57c^2 + 0.43c$    & 5.01E-04                           \\ \cline{2-5} 
                             & $0.52\pm 0.64$ & $0.22\pm 0.37$ & $-0.65c^3 + 0.65c$          & 6.14E-04                           \\ \hline
\multirow{10}{*}{$A\beta^+$} & $0.30\pm0.29$  & $0.16\pm 0.16$ & $-c^2+c$                    & 1.69E-03                           \\ \cline{2-5} 
                             & $0.30\pm 0.32$ & $0.18\pm 0.17$ & $-c^2+c$                    & 1.74E-03                           \\ \cline{2-5} 
                             & $0.29\pm 0.28$ & $0.18\pm 0.17$ & $-c^2+c$                    & 1.72E-03                           \\ \cline{2-5} 
                             & $0.31\pm 0.33$ & $0.16\pm 0.16$ & $-c^2+c$                    & 1.70E-03                           \\ \cline{2-5} 
                             & $0.29\pm 0.28$ & $0.15\pm 0.15$ & $-c^2+c$                    & 1.70E-03                           \\ \cline{2-5} 
                             & $0.31\pm 0.33$ & $0.15\pm 0.15$ & $-c^2+c$                    & 1.70E-03                           \\ \cline{2-5} 
                             & $0.30\pm 0.33$ & $0.17\pm 0.16$ & $0.11c^3 - 1.15c^2 + 1.03c$ & 1.71E-03                           \\ \cline{2-5} 
                             & $0.29\pm 0.28$ & $0.16\pm 0.16$ & $0.07c^3 - 1.08c^2 + 1.02c$ & 1.71E-03                           \\ \cline{2-5} 
                             & $0.29\pm 0.28$ & $0.16\pm 0.16$ & $-c^2+c$                    & 1.69E-03                           \\ \cline{2-5} 
                             & $0.29\pm 0.28$ & $0.15\pm 0.14$ & $0.23c^3 - 1.34c^2 + 1.11c$ & \textbf{1.68E-03} \\ \hline
\end{tabular}
\caption{\textbf{Identified $\alpha$, $\kappa$, $f_{sym}$ for the negative and positive groups in 10 independent simulations.} The lowest projection errors for each group are highlighted in bold.}
\label{tab:real_data_exp}
\end{table}

\subsubsection*{\normalfont{\textbf{Projection of tau concentration with uncertainty quantification}}}
\begin{figure}[!htbp]
\centering
\begin{subfigure}{\textwidth}
    \centering \includegraphics[width=0.99\textwidth]{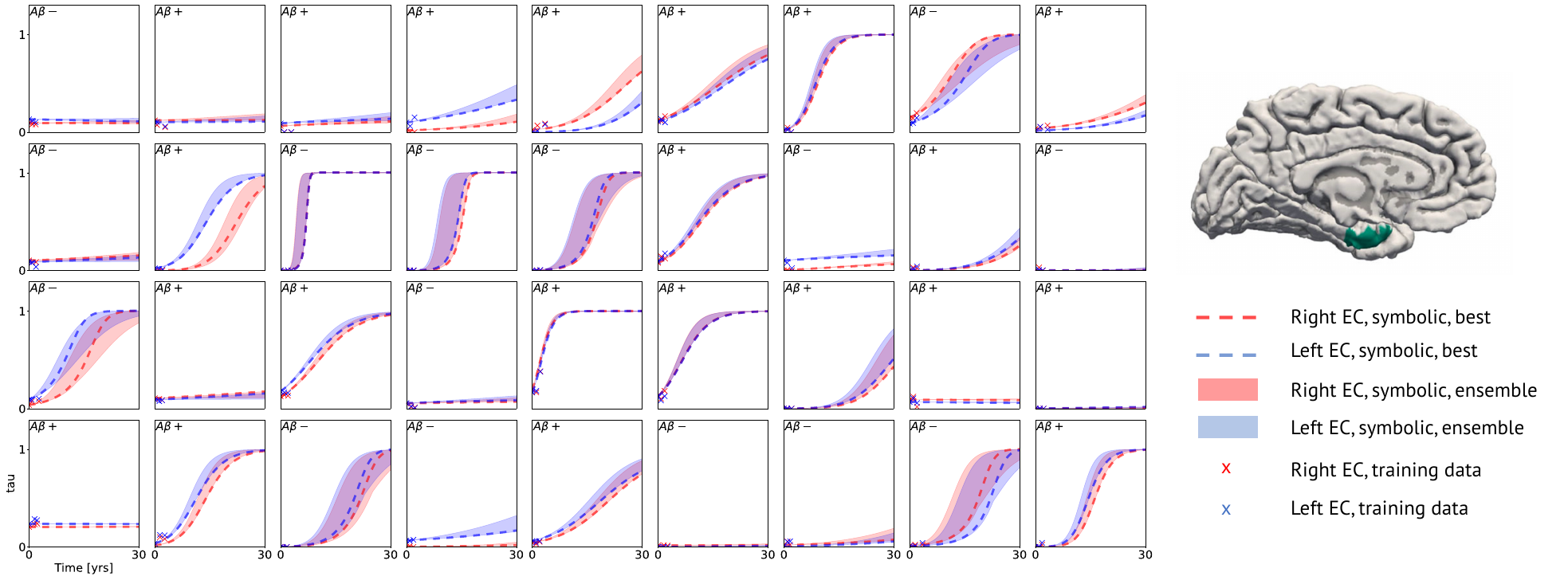}
\end{subfigure} \\
\begin{subfigure}{\textwidth}
    \centering \includegraphics[width=0.99\textwidth]{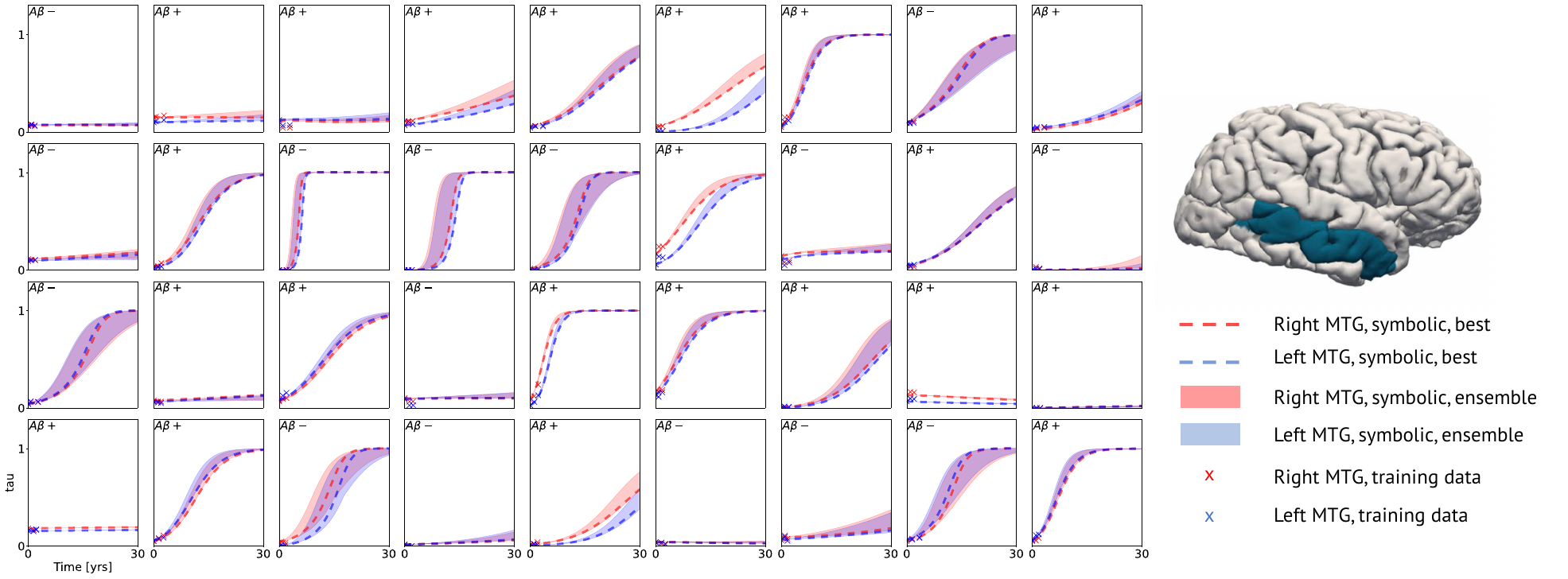}
\end{subfigure} \\
\begin{subfigure}{\textwidth}
    \centering \includegraphics[width=0.99\textwidth]{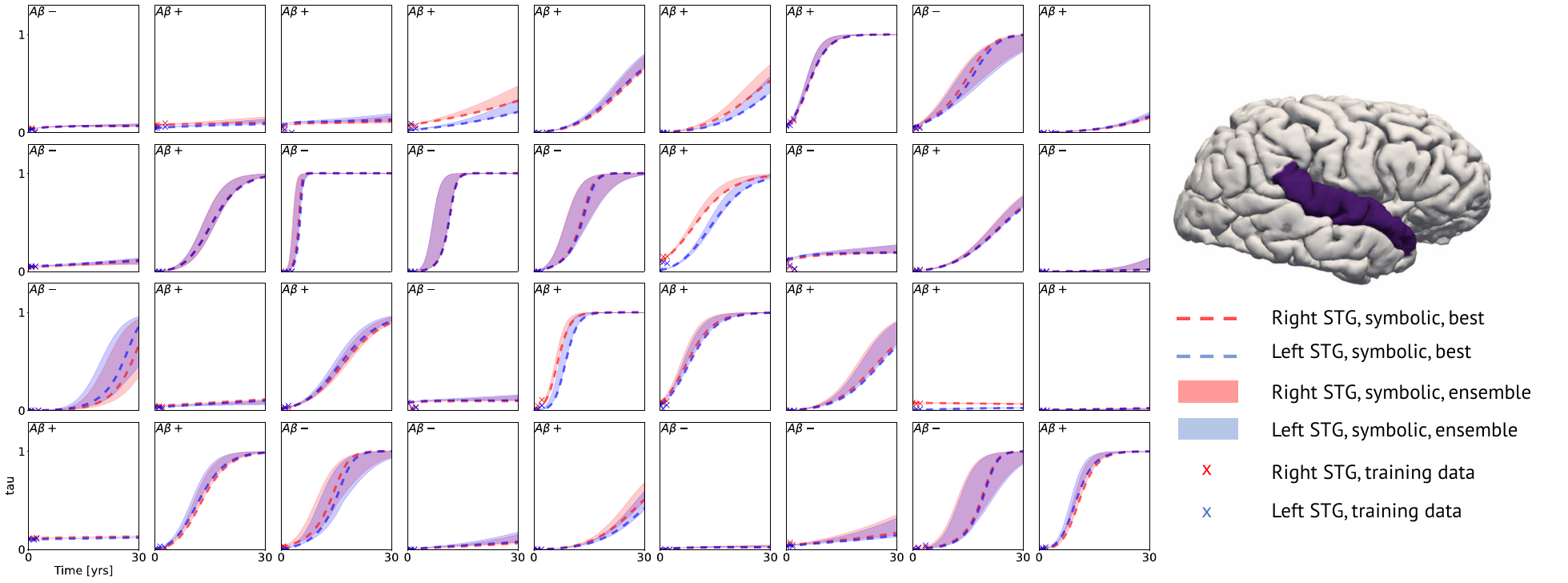}
\end{subfigure} 
\caption{\textbf{Projection of tau concentration in entorhinal cortex (EC), middle temporal gyrus (MTG) and superior temporal gyrus (STG) in 30 years based on real data in first three data points.} We plot the ensemble of predicted $c$ given by substituting $f_{sym}$ back to ODE of Eq.~\eqref{eq:discretized}. The lower and upper bound of the shaded region represents the minimum and maximum value of $c$ in all 10 simulations. We plot the predicted concentrations given by the model with lowest projection error in dashed lines.}
\label{fig:projection_real}
\end{figure}

We substitute the parameters inferred by PINN and reaction terms inferred from the symbolic regression model back to ODE of Eq.\eqref{eq:discretized} and extrapolate in time up to $t = 30$ for 36 subjects and three brain regions, similar to the synthetic data. Since ten independent simulations are conducted, we can propagate the uncertainty in the parameters ($\alpha$, $\kappa$) and functions ($f_{sym}^-$, $f_{sym}^+$) to quantify the uncertainty in the solution $c$. We obtain the extrapolated tau concentrations in each simulation, and plot their minimum and maximum values at each time step, as shown in Fig.~\ref{fig:projection_real}. The predicted concentrations given by the model with lowest projection error are plotted in dashed lines.

\section{Summary and discussion}\label{sec:4}

Tau protein is a key player in Alzheimer's disease, and misfolded tau proteins play a crucial role in disease progression and pathology. Protein misfolding and spreading across the brain of Alzheimer’s patients follows a characteristic stereotypical pattern that we can model with 
reaction-diffusion type equations. There is little controversy about the {\it{diffusion term}} of these equation: Since tau is an intracellular protein, the underlying assumption is that it spreads along axons within the brain’s connectome. However, the precise nature of the {\it{reaction term}} of these equations remains incompletely understood. 

Here, we capitalize on the recent developments in physics-informed deep learning and artificial intelligence to discover a mathematical model for the progression of Alzheimer’s disease from clinical data. Specifically, we use longitudinal tau positron emission tomography images from 46 individuals who are likely to develop Alzheimer’s disease and from 30 healthy controls. Their brain scans are publicly available through the Alzheimer’s Disease Neuroimaging Initiative database. By their very nature, positron emission tomography images have a high spatial resolution, while their temporal resolution is low and limited to only a few points in time. In essence, this temporal information is too sparse to infer protein misfolding dynamics from the medical images alone. This motivates the use of a physics-informed approach. 

Physics-informed neural networks or PINNs are a type of machine learning model that integrates our prior physics-based knowledge into the training process of neural networks. By incorporating the reaction-diffusion equation of protein misfolding into the loss function, PINNs 
satisfy the underlying physics by design, and improve the network’s accuracy and generalization capabilities. PINNs are particularly powerful for solving complex problems, when the underlying training data are sparse. While they seem well suited to simulate misfolded protein spreading across the brain from sparse clinical data, they will provide no insight into the functional form of the reaction dynamics.

Here, instead of using a stand-alone PINN simulation, we integrate {\it{physics informed neural networks}} and {\it{symbolic regression}} to discover a reaction-diffusion type partial differential equation for tau protein misfolding and spreading. Importantly, this is a two-step process in which the first step uses the PINN to learn a noninterpretable model from data and the second step uses symbolic regression to {\it{discover the best model and parameters}} to explain the reaction term inferred by the PINN. Importantly, our library of possible models contains a variety of popular engineering reaction terms and our parameters are interpretable by design.

We demonstrate the features our two-step approach in terms of synthetic and real data and discover the best model and parameters to explain tau imaging data from 46 individuals who are likely to develop Alzheimer’s disease and 30 healthy controls. Strikingly, our method discovers different misfolding models for the two groups, with a faster protein misfolding in the Alzheimer’s group, $f(c) = 0.23c^3 - 1.34c^2 + 1.11c$, than in the healthy control group, $f(c) = -c^3 +0.62c^2 + 0.39c$. We anticipate that our two-step modeling strategy generalizes well to other types of partial differential equations with various engineering applications. 

Taken together, our results suggest that PINNs, supplemented by symbolic regression, can discover a reaction-diffusion type model to explain misfolded tau protein concentrations in Alzheimer’s disease. Understanding the dynamics of tau protein misfolding and its propagation can provide insights into the mechanisms underlying Alzheimer's disease and potentially lead to the development of effective therapeutic interventions. We expect this study to be the starting point for a more comprehensive analysis to provide image-based technologies for early diagnosis, and ideally early treatment, of neurodegeneration in Alzheimer’s disease and possibly other misfolding-protein based neurodegenerative disorders.

\appendix
\section*{Appendix}
% \section{Pedagogical example: Discovering the K-O system}\label{appendix:A}
\section{Pedagogical example: Discovering the Kraichnan-Orszag system}\label{appendix:A}

For better understanding of the proposed integration of physics-informed neural networks (PINNs) and symbolic regression, in this section, we provide a pedagogical example that discovers the Kraichnan-Orszag system \cite{wan2006multi, zou2022neuraluq, chen2023leveraging} from data. The system is described by the following ODE system:
\begin{equation}\label{eq:ko}
    \begin{split}
        \frac{du_1}{dt} &= e^{-t/10}u_2u_3,\\
        \frac{du_2}{dt} &= u_1u_3,\\
        \frac{du_3}{dt} &= -2u_1u_2,
    \end{split}
\end{equation}
with initial conditions that are typically  drawn from a Gaussian distribution. Instead, here we fix the initial conditions as $u_1(0) = 1, u_2(0) = 0.8, u_3(0) = 0.5$, and assume partial knowledge of the dynamics. Specifically, we assume we know the right-hand side of the third equation to be a linear transformation of $u_1u_2$, and have zero information of the first two equations. Then we can rewrite the ODE system as follows:
\begin{equation}
    \begin{split}
        \frac{du_1}{dt} &= f_1(t, u_1, u_2, u_3),\\
        \frac{du_2}{dt} &= f_2(t, u_1, u_2, u_3),\\
        \frac{du_3}{dt} &= au_1u_2 + b,
    \end{split}
\end{equation}
where $a, b$ are unknown constants to be inferred and $f_1, f_2$ are unknown functions, whose analytic regressions we seek to discover. Here, we choose to use neural networks (NNs) with two hidden layers, each of which equipped with $50$ neurons, and hyperbolic tangent activation. Following the proposed method, we use one NN as the surrogate of $u$, which has $1$-dimensional input and $3$-dimensional output, and one NN to approximate concatenation of $f_1$ and $f_2$, which has $4$-dimensional input and $2$-dimensional output. The results are shown in Tables~\ref{tab:ko:1} and~\ref{tab:ko:2}. As we can see, both steps of our approach yield accurate inferences of $a, b$ and $f_1, f_2$. 

\begin{table}[ht]
    \footnotesize
    \centering
    \begin{tabular}{|c|c|c|}
        \hline
         & a & b \\
         \hline
         Inference & -1.9975 & 0.0002 \\
         \hline
         Exact & -2 & 0 \\
         \hline
    \end{tabular}
    \caption{Inference of unknown constants using PINNs.}
    \label{tab:ko:1}
\end{table}

\begin{table}[ht]
    \footnotesize
    \centering
    \begin{tabular}{|c|c|c|c|}
         \hline
         $f_1$ & $u_2 u_3 e^{-0.0986t}$ & $0.6694 u_2 u_3$ & $u_2u_3$ \\
         \hline
         Score & 3.8855 & 0.6695 & 0.5910 \\
         \hline
         \hline
         $f_2$ & $u_1u_3$ & $u_1u_3 + \frac{0.0014}{t+0.0314}$ & $u_1u_3 - 0.0020 + \frac{0.0017}{t + 0.0379}$ \\
         \hline
         Scores & 2.9805 & 0.5552 & 0.0617 \\
         \hline
    \end{tabular}
    \caption{Identifications of the dynamics of $u_1$ and $u_2$ using PINNs and symbolic regression. Top 3 identifications, in terms of the score defined in Section \ref{subsec:symbolic}, are shown.}
    \label{tab:ko:2}
\end{table}

We reiterate that $a$ and $b$ are jointly inferred in the PINNs step while the identification of $f_1$ and $f_2$ is done separately in the symbolic regression step. We also note that results from symbolic regression may be sensitive to the metrics of model selection, to the candidates of unary and binary operators, and to the complexities of the operators. In this example, the metric is the score defined in \cite{cranmer2020discovering} and described in Section \ref{subsec:symbolic}, the binary operators are chosen to be addition and multiplication, the unary operators are identity, $\sin$, $\cos$, exponential and reciprocal, and the complexities of unary operators are set to 1, 3, 3, 3, 3, respectively.

\section{Effect of sample size on inference quality}
\label{appendix:sample_size}

\begin{figure}[!htbp]
\includegraphics[width=\textwidth]{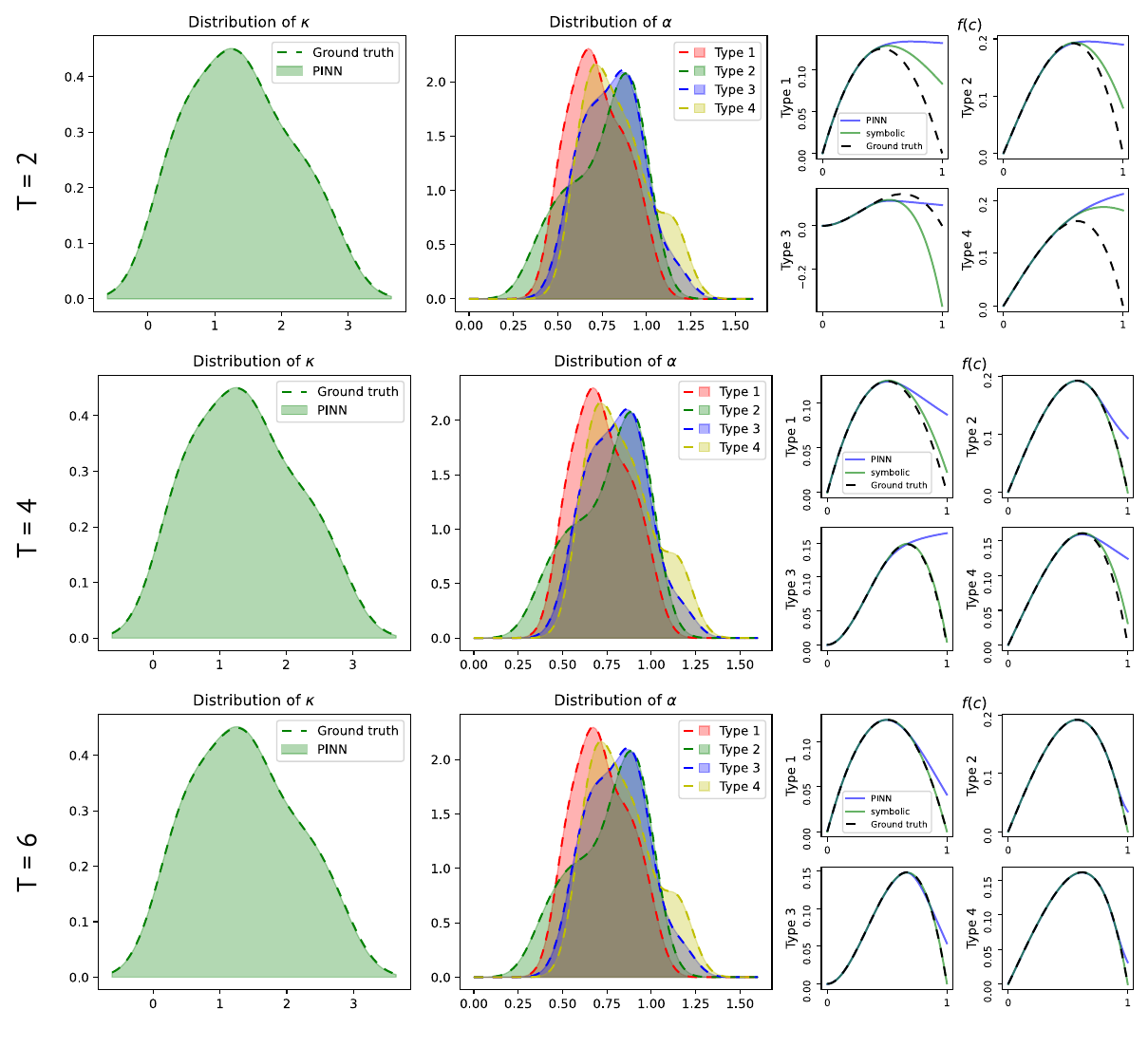}
\caption{\textbf{PINN inference results when $T = 2, 4, 6$, without fixed boundary conditions for $f$.} While the learned parameters $\kappa$ and $\alpha$ perform well in all instances, the parametric function $f_{\phi}$ fails to align with $f$ in regions where there is no available data for $c$.}
\label{fig:params_appendix}
\end{figure}

In the simulated dataset, we are provided with training data $c(t)$ at $t = 0, 1, \cdots, T$. In the primary exposition, we have determined that a value of $T=2$ suffices for the purpose of training, inference, and projection. This conclusion is, in part, attributable to the additional physical constraint that we have imposed, namely, $f(0) = f(1) = 0$. In the absence of these constraints, our investigation has revealed that a greater amount of data is required to optimize the model effectively. To illustrate this finding, we have conducted three simulations with $T=2,4,6$ and have plotted the inference results obtained using PINNs and the symbolic model, as depicted in Figure~\ref{fig:params_appendix}. It is noteworthy that in all cases, the parameters $\kappa$ and $\alpha$ are inferred correctly; yet, the parametric function $f_{\phi}$ does not match $f$ in areas where no data for $c$ is available. This is particularly pronounced for larger concentration $c$ values, which correspond to subjects with tau concentration levels close to 0 in the initial stages. The rightmost column of Figure~\ref{fig:params_appendix} shows that the inferred $f(c)$ does not agree well with the ground truth. Interestingly, when $T=4$ and $T=6$, the symbolic regression model $f_{sym}$ discovered from the neural network $f_{\phi}$ outperforms $f_{\phi}$ in approximating the underlying target. One possible explanation for this phenomenon is that $f_{sym}$ is trained with the concentration $c$ from the training set of $f_{\phi}$, which means that the region that corresponds to the incorrect inference by $f_{\phi}$ is not fed into the symbolic regression model. As a result, the underlying inductive bias directs $f_{sym}$ towards the correct solution automatically. Finally, we note that Table~\ref{tab:simulated_data} demonstrates that, to obtain the correct symbolic model when there are no constraints on the boundary conditions of $f$, data up to $T=6$ is required.

\begin{table}[!htbp]
\footnotesize
\centering
\begin{tabular}{|l|l|l|l|l|}
\hline
& Group 1 & Group 2 & Group 3 & Group 4\\ \hline
$T = 2$ & $2.88ce^{-1.05e^c}$ & $c(-1+2e^{-0.55c^2})$ & $c^2(1.76-0.78e^c)$ & $0.37ce^{-0.71c^2}$ \\ \hline
$T = 4$ & $c(-1.16c+0.12e^c+0.9)$ & $c(1-c^2)$ & $c^2(1-c)$ & $c(0.51c-0.49e^c+0.86)$   \\ \hline
$T = 6$ & $c(1-c)$ & $c(1-c^2)$ & $c^2(1-c)$ & 
$c(0.37-0.37c)e^c$\\ \hline
\end{tabular}
\hfill
\caption{\textbf{Inferred reaction terms when $T = 2, 4, 6$, without fixed boundary conditions for $f$.} When boundary conditions for $f$ are not fixed, more data are required to guarantee the correct inference result by symbolic regression model. The more data we are provided, the better inference quality will be achieved. To get $f_{sym}$ correct in 4 groups, we need data up to $T = 6$.}
\label{tab:simulated_data}
\end{table}
\newpage

\section*{Acknowledgements}
We would like to thank Dr. Amelie Sch{\"a}fer and Dr. Kevin Linka for helpful discussions. ZZ, ZZ and GEK were supported by the DOE SEA-CROGS project (DE-SC0023191), the MURI-AFOSR FA9550-20-1-0358 project, and the ONR Vannevar Bush Faculty Fellowship (N00014-22-1-2795). EK was supported by 
the NSF CMMI grant 2320933 Automated Model Discovery for Soft Matter.

\bibliographystyle{unsrt}
\bibliography{reference}

% \section*{Data availability}
% The data used in this study are available from the corresponding authors upon reasonable request.

% \section*{Code availability}
% The codes to produce computational results in this work will be made available to public upon acceptance of the manuscript.

% \section*{Author contributions}

% EK and GEK designed the study and supervised this work. ZZ and ZZ developed, implemented, and tested the method. Z. Zhang produced the computational results. EK provided the experimental setup and the data. All authors wrote and revised the manuscript.

% \section*{Competing interests}
% Karniadakis provides technical advice on directions in machine learning to Anailytica, a private startup company developing AI software products for engineering. He has a very small equity for his work. The rest of the authors declare no competing interests.

\end{document}